\documentclass[conference]{IEEEtran}
\usepackage{times}

\usepackage[numbers]{natbib}
\usepackage{multicol}
\usepackage[bookmarks=true]{hyperref}
\usepackage[utf8]{inputenc}
\usepackage{pdfpages}
\usepackage{lscape}
\usepackage[letterpaper, margin=0.75in]{geometry}
\usepackage{textcomp}
\usepackage{graphicx}
\usepackage[version=4]{mhchem}
\usepackage{siunitx}
\usepackage{longtable,tabularx}
\usepackage{accents}
\usepackage{comment}
\usepackage{caption}
\usepackage{color}
\usepackage{booktabs}
\usepackage{tabu}
\usepackage{soul}
\usepackage{fancyhdr}
\usepackage{subfigure}
\usepackage{wrapfig}
\usepackage{amsmath}
\usepackage{amssymb}
\usepackage{colortbl}

\DeclareMathOperator{\subjto}{subject\,to}
\DeclareMathOperator{\minimize}{minimize}

\DeclareMathOperator{\trace}{tr}

\DeclareMathOperator{\diag}{diag}
\newcommand{\set}[1]{\mathcal{#1}}

\newcommand{\mat}[1]{\mathbf{#1}}
\renewcommand{\Vec}[1]{\mathbf{#1}}

\usepackage{tikz}
\usepackage{siunitx} 
\usetikzlibrary{arrows, arrows.meta}
\usetikzlibrary{graphs}
\usetikzlibrary{bayesnet}
\usetikzlibrary{backgrounds}

\usepackage{standalone}

\tikzstyle{target}=[draw,fill=yellow!50,circle,minimum size=16pt,inner sep=0pt]

\tikzstyle{output}=[draw,fill=blue!50,circle,minimum size=16pt,inner sep=0pt]
\tikzstyle{bias}=[draw,fill=gray!50,circle,minimum size=20pt,inner sep=2pt]
\tikzstyle{arrow}=[arrows={{Latex[scale=0.5]}-}, thick]  
\tikzstyle{box}=[rectangle, draw=black!100] 

\tikzset{
    between/.style args={#1 and #2}{
         at = ($(#1)!0.5!(#2)$)
    }
}

\begin{document}

\title{Exploiting Structure for Optimal Multi-Agent Bayesian Decentralized Estimation}



%
\author{\authorblockN{Christopher Funk\authorrefmark{1}, 
Ofer Dagan\authorrefmark{2},
Benjamin Noack\authorrefmark{1} 
and Nisar R. Ahmed\authorrefmark{2}}
\authorblockA{\authorrefmark{1}Institute for Intelligent Cooperating Systems,\\ Otto von Guericke University Magdeburg, Magdeburg, Germany\\ 
Email: christopher.funk@ovgu.de,benjamin.noack@ieee.org
}
\authorblockA{\authorrefmark{3}Smead Aerospace Engineering Sciences Department,\\ University of Colorado Boulder, Boulder, CO 80309 USA\\
Email: \{ofer.dagan, nisar.ahmed\}@colorado.edu
}
}

\maketitle

\begin{abstract}
A key challenge in Bayesian decentralized data fusion is the `rumor propagation' or `double counting' phenomenon, where previously sent data circulates back to its sender.
It is often addressed by approximate methods like covariance intersection (CI) which takes a weighted average of the estimates to compute the bound.
The problem is that this bound is not tight, i.e. the estimate is often over-conservative.
In this paper, we show that by exploiting the probabilistic independence structure in multi-agent decentralized fusion problems a tighter bound can be found using (i) an expansion to the CI algorithm that uses multiple (non-monolithic) weighting factors instead of one (monolithic) factor in the original CI and (ii) a general optimization scheme that is able to compute optimal bounds and fully exploit an arbitrary dependency structure.
We compare our methods and show that on a simple problem, they converge to the same solution. 
We then test our new non-monolithic CI algorithm on a large-scale target tracking simulation and show that it achieves a tighter bound and a more accurate estimate compared to the original monolithic CI.

\end{abstract}

\IEEEpeerreviewmaketitle

\section{Introduction}



\begin{figure}[tb]
    \centering
   \includegraphics[width=0.35\textwidth]{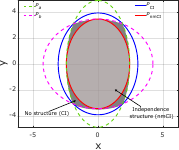}    
    \caption{Intuitive example of how exploiting independence structure leads to a tighter bound. With no structure possible fused covariances occupy the intersection of $\mat P_a$ and $\mat P_b$ (dark gray). With structure, all possible results are limited to the light gray area, thus allowing nmCI to find a tighter bound. }
    \label{fig:fused_ellipsoids}
    \vspace{-0.3in}
\end{figure}

In many multi-agent applications such as multi-target tracking \cite{dagan_exact_2023}, \cite{whitacre_decentralized_2011}, simultaneous localization and mapping (SLAM) \cite{cunningham_ddf-sam_2013}, and cooperative localization \cite{carrillo-arce_decentralized_2013}, \cite{loefgren_scalable_2019}, 
Bayesian decentralized data fusion (DDF) is used for peer-to-peer fusion of estimates. 
A key challenge in DDF is the so-called `rumor propagation', where due to common data between the agents, their estimates might be correlated.
Since tracking the common data requires pedigree tracking \cite{martin_distributed_2005}, which might be cumbersome in large networks, or limits the network topology \cite{grime_data_1994}, methods that try to bound the true uncertainty in the face of unknown correlation between the estimates gained popularity.  
One of the most commonly used methods for two probability distribution functions (pdf), represented by their first two moments (mean and covariance), is covariance intersection (CI) \cite{julier_non-divergent_1997}.
For intuition consider Fig. \ref{fig:fused_ellipsoids}, the true fused covariance must be enclosed by the intersection of the two prior covariances $\mat P_a$ and $\mat P_b$ (dark gray). 
CI optimizes between all covariances that pass through the intersection points of $\mat P_a$ and $\mat P_b$ and finds the optimal bound on the true fused covariance \cite{reinhardt_minimum_2015}.

However, in its most basic form \cite{julier_non-divergent_1997}, it does not make any use of the underlying independence structure between variables, thus the bound achieved by CI might not be tight.
For example, when the variables represent two independent states,
the true fused covariance is limited to certain areas of the intersection, as shown in \ref{fig:fused_ellipsoids} (light gray).
In this paper, we show that this independence structure can be exploited to find a tighter bound using (i) an expansion to the CI algorithm -- the non-monolithic CI (nmCI) and (ii) a more general scheme that is able to compute optimal bounds and fully exploit an \textit{arbitrary} dependency structure.
The ability to compute optimal bounds in the latter case, even if not tractable in real-time, establishes a baseline of performance for other fusion approaches and provides a tool to empirically investigate the optimality and conservativeness of new fusion approaches. 
Previous work on this topic may be found in \cite{forsling_conservative_2022}, which uses the same optimization formalism as this work. 
However, no algorithmic details are divulged and optimal solutions cannot be guaranteed. 
In comparison, this work may be seen as the first approach to compute guaranteed asymptotically optimal solutions for the general case.

\label{sec:intro}

\section{Problem Statement}
Consider a decentralized network of $n_a$ autonomous Bayesian agents, tasked with jointly monitoring a set $\chi$ of random variables (rvs).
Each agent $a$ recursively updates its local prior pdf $p(\chi)$ with (i) independent sensor measurements, described by $p(z_k^a|\chi_k^a)$, the likelihood of observing $z_k^a$ conditioned on the subset of rvs $\chi_k^a$ at time step $k$, and (ii) data set $Z^b_k$, received from a neighboring agent $b\in N_a^a$ via the peer-to-peer distributed variant of Bayes' rule \cite{chong_distributed_1983}, 
\begin{equation}
    p_f(\chi|Z_k^a\cup Z_k^b)\propto \frac{p^a(\chi|Z^{a}_{k})p^b(\chi|Z^{b}_{k})}{p^{ab}_c(\chi|Z_{k}^{a}\cap Z_{k}^{b})}.
    \label{eq:bayesCF}
\end{equation}
Here $p^{ab}_c(\chi|Z_{k}^{a}\cap Z_{k}^{b})$ denotes the pdf over $\chi$ given the data common to agents $a$ and $b$, and needs to be removed in order to prevent `double counting' of previously shared data.

The main challenge in DDF is to account for the denominator in (\ref{eq:bayesCF}). 
While the denominator can be tracked explicitly, as discussed in the introduction,
in this paper, we consider the problem where the dependency in the data, i.e. the common data between the agents ($Z_{k}^{a}\cap Z_{k}^{b}$) is unknown.
We assume that some independence structure between rvs exists, e.g., that the northern and eastern states of a tracked target are independent of each other, and aim to 
find an optimal (according to some measure) approximation $\bar{p}_f(\chi|Z_k^a\cup Z_k^b)$ that is a conservative approximation of the true fused posterior pdf $p_f$ in (\ref{eq:bayesCF}),
\begin{equation}
    \bar{p}_f(\chi|Z_k^a\cup Z_k^b) \succeq p_f(\chi|Z_k^a\cup Z_k^b).
    \label{conservativePDF}
\end{equation}
Where here we consider 
$\bar{p}_f(\cdot)$ to be a conservative approximation of $p_f(\cdot)$ if $\bar{\mat P}_f-\mat P_f\succeq 0$,
where $\bar{\mat P}_f$ and $\mat P_f$ are the covariances of $\bar{p}_f(\cdot)$ and $p_f(\cdot)$, respectively.

Assume then that $p_f(\cdot)$ is described by its first two moments (mean and covariance), and let $\Vec \chi_a$, $\Vec \chi_b$ denote two means/point estimates with associated covariances $\mat P_a$, $\mat P_b$ and sparse correlation $\mat P_{ab}$. 
The sparsity structure is implied by the inherent probabilistic properties of the system being estimated.
In the following, we derive a robust optimization problem to simultaneously determine the optimal gains $\mat K_a$ and $\mat K_b$ of the linear fusion rule $\Vec \chi_f=\mat K_a\Vec \chi_a+\mat K_b\Vec \chi_b,$ where $\Vec \chi_f$ is the fusion result, and a minimal upper bound $\bar{\mat P}_f$ on the fusion result error covariance 
\begin{equation*}
    \mat P_f=\mat K_a \mat P_{aa}\mat K_a^\top+\mat K_a \mat P_{ab} \mat K_b^\top+\mat K_b\mat P_{ab}^\top\mat K_a^\top+\mat K_b\mat P_{bb}\mat K_b^\top,
\end{equation*}
which is unknown, due to $\mat P_{ab}$ being unknown.

We begin by deriving a sufficient condition for $\bar{\mat P}_f$ to be an upper bound on $\mat P_f$, i.e., $\bar{\mat P}_f\succeq\mat P_f$. To do so, note that $\mat P_{ab}$ cannot take arbitrary values, but only those that result in the joint covariance $\mat P=\begin{bmatrix}\mat P_a & \mat P_{ab}\\\mat P_{ab}^\top & \mat P_b\end{bmatrix}$ being positive definite. Hence, we require that $\bar{\mat P}_f\succeq\mat P_f$ for any such $\mat P_{ab}$. In particular, this implies that $\bar{\mat P}_f\succeq\mat P_f$ holds for the true (but unknown) value of $\mat P_{ab}$.
Next, we introduce the condition $\mat K_a+\mat K_b=\mat I$ to ensure that $\Vec \chi_f$ is unbiased, should the means / point estimates $\Vec \chi_a$ and $\Vec \chi_b$ be unbiased.
Finally, together with the objective function $\trace(\bar{\mat P}_f)$, which is strictly matrix increasing, i.e., $\mat A\prec\mat B\implies\trace(\mat A)<\trace(\mat B)$, this gives us the robust optimization problem
\begin{align}
    \underset{\mat K_a,\mat K_b,\bar{\mat P}_f}{\minimize} &\quad \trace(\bar{\mat P}_f)\label{eq:robust_sdp}\\
    \subjto &\quad \mat K_a+\mat K_b=\mat I\nonumber\\
    &\quad \bar{\mat P}_f\succeq \mat K_a \mat P_a\mat K_a^\top+\mat K_a \mat P_{ab}' \mat K_b^\top\quad\ \forall\mat P_{ab}'\in\set U\nonumber\\
    &\quad\phantom{\bar{\mat P}_f\succeq} +\mat K_b\mat P_{ab}'^\top\mat K_a^\top+\mat K_b\mat P_b\mat K_b^\top\nonumber,
\end{align}
that minimizes the conservativeness of the bound. This robust optimization problem is equivalent to that presented in \cite{forsling_conservative_2022}. In the above problem, the set $\set U$ in the constraint is given by 
\begin{equation*}
    \set U=\left\{\mat P_{ab}'\middle|\begin{bmatrix}
        \mat P_a & \mat P_{ab}'\\
        \mat P_{ab}'^\top & \mat P_b
    \end{bmatrix}\succ\mat 0, [\mat P_{ab}']_{ij}=0,(i,j)\in\set I\right\},
\end{equation*}
where $\set I$ is the index set of the known zero elements of $\mat P_{ab}$.
Based on \eqref{eq:robust_sdp} it becomes clear that the more elements of $\mat P_{ab}$ are zero, the tighter the achievable bound $\bar{\mat P}$ becomes. This is due to $\set U$ getting smaller as more elements of $\mat P_{ab}$ are known. Smaller $\set U$ correspond to a larger feasible set, which allows for smaller optimal objective values. When $\mat P_{ab}$ is completely unknown, the above problem's solution simplifies to the CI solution \cite{julier_non-divergent_1997}, \cite{reinhardt_minimum_2015}.
In the following, we are interested in (approximately) solving the general problem given by \eqref{eq:robust_sdp}, which results in tighter bounds, as shown in Fig. \ref{fig:fused_ellipsoids}.
\label{sec:probStatement}

\section{Technical Approach}
\label{sec:Technical_approach}
To describe our technical approach consider a simple $2D$ tracking example, where $n_a=2$ autonomous agents $a$ and $b$ are tracking the $x$ and $y$ position of $n_t=1$ dynamic target, thus $\chi=[x,y]^T$.
Assume the agents have perfect knowledge of their ego position and take noisy measurements $y_k$ of the target position.
In this case, assuming that the dynamics of the target are separated along the $x$ and $y$ directions, then the rvs describing the position of the target $x$ and $y$, are independent, thus $p(\chi)=p(x,y)=p(x)\cdot p(y)$.
The key insight here is that by leveraging the independence structure of the problem we can separate the data fusion problem (\ref{eq:bayesCF}) into two independent fusion problems, for $x$ and $y$.  
\begin{equation}
    p_f(\chi|Z_k^a\cup Z_k^b)\propto \frac{p^a(x|Z^{a}_{k})p^b(x|Z^{b}_{k})}{p^{ab}_c(x|Z_{k}^{a}\cap Z_{k}^{b})}\cdot \frac{p^a(y|Z^{a}_{k})p^b(y|Z^{b}_{k})}{p^{ab}_c(y|Z_{k}^{a}\cap Z_{k}^{b})}.
    \label{eq:nmBayes}
\end{equation}

In the following sections, we exploit the problem independence structure to (i) extend the (monolithic) CI fusion rule to a new, less conservative, \emph{non-monolithic CI} (nmCI) fusion rule, and (ii) approximate the optimal conservative fusion result using semidefinite programming.

\subsection{Non-Monolithic Fusion}
We start our derivation of the new fusion rule with the geometric mean density (GMD) \cite{hurley_information_2002} \cite{bailey_conservative_2012}, which is a generalization of CI to pdf fusion \cite{mahler_optimal/robust_2000}, 
\begin{equation}
    p_f(\chi)\propto p^a(\chi)^\omega\cdot p^b(\chi)^{1-\omega}, \ \ \ 0\leq\omega\leq 1.
    \label{eq:GMD}
\end{equation}
Here $\omega$ is a scalar weighting constant, chosen according to a desired cost metric \cite{ahmed_fast_2012}, and the dependency on the data $Z$ is omitted for brevity. 
Assuming, as described above, that the random state vector $\chi$ can be divided into two independent random states $x$ and $y$, we rewrite (\ref{eq:GMD}) as
\begin{equation}
\begin{split}
    p_f(\chi)\propto p^a(x)^{\omega_x}\cdot &p^b(x)^{1-\omega_x}\cdot p^a(y)^{\omega_y}\cdot p^b(y)^{1-\omega_y}, \\
    &0\leq\omega_x, \omega_y\leq 1.
    \label{eq:nmGMD}
    \end{split}
\end{equation}
Now the weighting constant $\vec{\omega}=[\omega_x, \omega_y]$ is no longer a scalar, but a non-monolithic vector of weights. 
We dub this fusion rule nmGMD.

From here we focus our attention on Gaussian pdfs and cases where the non-Gaussian pdf is expressed using the first two moments.
It has been shown \cite{hurley_information_2002} that for Gaussian pdfs, the GMD fusion is equivalent to the CI fusion rule,
\begin{equation}
    \begin{split}
        \mat P^{-1}_f&=\omega \mat P_a^{-1}+(1-\omega) \mat P_b^{-1},\\
        \mat P^{-1}_f\mu_f&=\omega \mat P_a^{-1}\mu_a+(1-\omega) \mat P_b^{-1}\mu_b,
    \end{split}
    \label{eq:CI}
\end{equation}
where $\{\mu_a, \mat P_a\}$ ($\{\mu_b, \mat P_b\}$) are the estimate mean and covariance, respectively, of agent $a$ ($b$). 
In the presence of an unknown degree of correlation between the estimates of agents $a$ and $b$, i.e. when $\mat P_{ab}$ is unknown, the CI rule produces the optimal fusion result \cite{reinhardt_minimum_2015}.
However, when the parts of the state are independent, we can deduce that the estimates must also be independent or uncorrelated. 
Consider again the example given at the beginning of Sec. \ref{sec:Technical_approach}, if $x$ and $y$ are independent rvs, then the correlation between agent $a$'s estimate of $x$ and agent $b$'s estimate of y, must be $0$.
The independence structure of the underlying estimation problem implies the following cross-covariance matrix, $\mat P_{ab}=\diag(\sigma^x_{ab},\ \sigma^y_{ab})$, 
where $\sigma^x_{ab} \ (\sigma^y_{ab})$ are the cross-correlations between the $x \ (y)$ estimates held by $a$ and $b$.


\subsection{Approximately Optimal Fusion Gains and Bounds}
Unfortunately, robust optimization problems are often not computationally tractable \cite{ben-tal_robust_2002}. Here, we make \eqref{eq:robust_sdp} manageable by approximating the uncountable uncertainty set $\set U$ by a random $n$-element subset $\set U_n=\{\mat P_{ab,1}',\ldots,\mat P_{ab,n}'\}\subset\set U$. 
This simplification of \eqref{eq:robust_sdp} comes at the risk of possibly non-conservative optimal solutions, as $\bar{\mat P}_f$ is no longer enforced to be conservative for every possible $\mat P_{ab}$. However, it seems intuitive that for $n\rightarrow\infty$ the optimal solution of the simplified problem approaches that of \eqref{eq:robust_sdp}, which is conservative. This conjecture and approaches to quantify the non-conservativeness are the subjects of ongoing research.

Two components are required to make the approach sketched above work: (i) a method to solve the simplified problem efficiently and (ii) a method to randomly draw $\mat P_{ab,i}'$ from a distribution with support given by $\set U$. The condition on the support guarantees that every subset of $\set U$ will eventually be drawn from.
The first component follows from applying the Schur complement condition for positive semidefiniteness
to the positive semidefinite constraints in \eqref{eq:robust_sdp}, each of which can be written in the required form $\bar{\mat P}_f-
    \mat K
    \mat P_i'
    \mat K^\top
    \succeq \mat 0$
where $\mat K=[\mat K_a, \mat K_b]$ and 
$\mat P_i'=\begin{bmatrix}
        \mat P_a & \mat P_{ab,i}'\\
        \mat P_{ab,i}'^\top & \mat P_b
    \end{bmatrix}$.
This gives the equivalent optimization problem
\begin{align}
    \underset{\mat K_a,\mat K_b,\bar{\mat P}_f}{\minimize} &\quad \trace(\bar{\mat P}_f)\label{eq:sampled_sdp}\\
    \subjto &\quad \mat K_a+\mat K_b=\mat I\nonumber\\
    &\quad \begin{bmatrix}
        \bar{\mat P}_f & \mat K\\
        \mat K^\top & \mat P_i'^{-1}
    \end{bmatrix}\succeq\mat 0\quad\ i=1,\ldots,n,\nonumber
\end{align}
which is a standard semidefinite program and can be solved in polynomial time by any off-the-shelf SDP solver.
For the second component, we use rejection sampling. Because $\set U$ is unbounded, we do not sample covariance matrix cross-terms $\mat P_{ab,i}'$ from $\set U$ directly, but instead sample equivalent correlation matrix cross-terms $\mat C_{ab,i}'$.
This has the advantage that the elements of the $\mat C_{ab,i}'$ are in $[-1,1]$. 
Hence, the rejection sampling process is as follows:
(i) Sample $\mat C_{ab,i}'$ from the proposal distribution. This corresponds to drawing each element of $\mat C_{ab,i}'$ that is not fixed to zero by the sparsity pattern of $\mat P_{ab}$ uniformly from $[-1,1]$.
(ii) Accept $\mat C_{ab,i}'$ if the resulting joint correlation matrix is positive definite. Reject otherwise and repeat (i) and (ii).
(iii) If $\mat C_{ab,i}'$ was accepted, compute $\mat P_i'$ from the joint correlation matrix that results from $\mat C_{ab,i}'$.
Once $n$ samples have been collected, \eqref{eq:sampled_sdp} can be constructed and subsequently solved.

\section{Results and Discussion}
\label{sec:Results_Dicussion}

In this section, we first compare the optimization solution to the nmCI solution for a $2D$ problem and show that in this case, the nmCI solution is the \emph{optimal} solution.
Then we test the nmCI and compare it to the monolithic CI algorithm on a larger-scale problem,  with 16 agents, 20 targets, and $112$ states target tracking simulation. 

\subsection{Comparison: nmCI and Optimization Approaches}
\begin{figure}
    \centering
    \includegraphics[width=0.95\columnwidth]{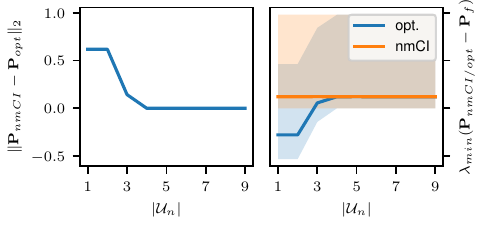}
    \vspace{-0.1in}
    \caption{Left: Deviation of nmCI bound to optimized bound. Right: Median, maximum/minimum of the smallest eigenvalue of difference between computed bound and actual fused covariance.}
    \label{fig:nmci_opt_results}
    \vspace{-0.2in}
\end{figure}

To compare nmCI and the optimization-based approach we consider the following simple scenario with a 2D state vector and two state estimates. The covariances of the estimates are given by $\mat P_a=\diag(3,1)$, $\mat P_b=\diag(1,4)$, and their correlation is assumed to be of the form $\mat P_{ab}=\diag(\cdot,\cdot)$. This occurs when the state consists, e.g., of the $x$ and $y$ position of a target, where the positions are independent due to the 
probabilistic model describing the target's motion and the measurements taken of it. 
Here, the nmCI solution is given by $\mat P_{nmCI}=\diag(1,1)$, corresponding to $\vec{\omega}=[0, 1]$.
We now apply the optimization approach to the above problem for different cardinalities of the set $\set U_n$. For each cardinality we perform $1000$ Monte Carlo (MC) runs, where the true $\mat P_{ab}$ is randomly sampled according to the procedure used in the optimization approach. The deviations $\|\mat P_{nmCI}-\mat P_{opt}\|_2$ between the nmCI result and optimization result in $\mat P_{opt}$ and the medians, maximums, and minimums of the minimum eigenvalues of $\mat P_{nmCI}-\mat P_f$ and $\mat P_{opt}-\mat P_f$ over the MC runs are shown in Fig.\ref{fig:nmci_opt_results}.
As can be seen, the nmCI and optimized solution do not agree for small $|\set U_n|$ but converge to each other for larger values.
While nmCI is always conservative, the optimization method results in non-conservativeness for too small $|\set U_n|$, and interestingly, eventually becomes conservative at some finite $|\set U_n|$. 
Because the optimized bound is optimal for \eqref{eq:sampled_sdp} and identical to the guaranteed conservative nmCI bound for large enough $|\set U_n|$, it is optimal for \eqref{eq:robust_sdp} when $|\set U_n|$ is large enough. Conversely, the nmCI solution must be optimal for \eqref{eq:robust_sdp}.

\subsection{Simulation: Multi-Agent Multi-Target Tracking}
\begin{figure}[tb]
    \centering
    \includegraphics[width=0.45\textwidth]{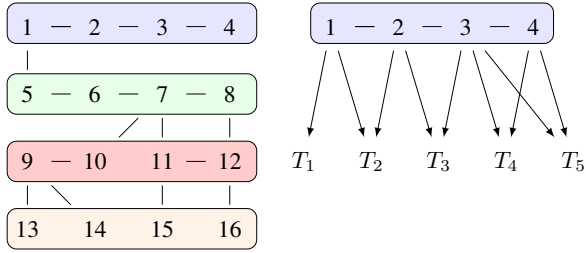}
    \caption{Left: Undirected and cyclic network topology, split into 4 groups of 4 agents. Right: Target ($T_t$) tracking assignments to robots 1--4, other groups follow the same pattern. }
    \label{fig:net_topology}
    \vspace{-0.25in}    
\end{figure}

To validate the algorithms and demonstrate the advantage of the new nmCI algorithm over the original CI algorithm, we perform a 16-agent, 20-target localization and tracking simulation.
In this scenario, as described by Fig. \ref{fig:net_topology}, the agents are split into 4 groups of 4 agents, where each group monitors 5 (mutually exclusive) targets out of the 20. 
Note that this means that the set of states monitored by one group is independent of the set of states monitored by another group. 
Assume each tracking agent $a\ =1,...,16$ has perfect self-position knowledge but with a constant agent-target relative position measurement bias vector in the $x$ and $y$ directions $s^a=[b^a_{x},b^a_{y}]^T$. 
In every time step $k$, each agent $a$ takes two measurements $z_k^{a,t}$ to target $t$, and $m^a_k$ to a landmark, 
\begin{align}
    \begin{split}
        z^{a,t}_{k} &= H\chi^t+s^a+v^{a,1}_k, \ \ v^{a,1}_k \sim \mathcal{N}(0,R^{a,1}),  \\
        m^a_{k} &= s^a+v^{a,2}_k, \ \ v^{a,2}_k \sim \mathcal{N}(0,R^{a,2}),
    \end{split}
    \label{eq:measModel}
\end{align}
where $\chi^t={[x^t,\dot{x}^t, y^t, \dot{y}^t]}^T$ is the $x$ and $y$ position and velocity of target $t$, and $H$ is the measurement matrix.





\begin{figure}[tb]
    \centering
    \includegraphics[width=0.3\textwidth]{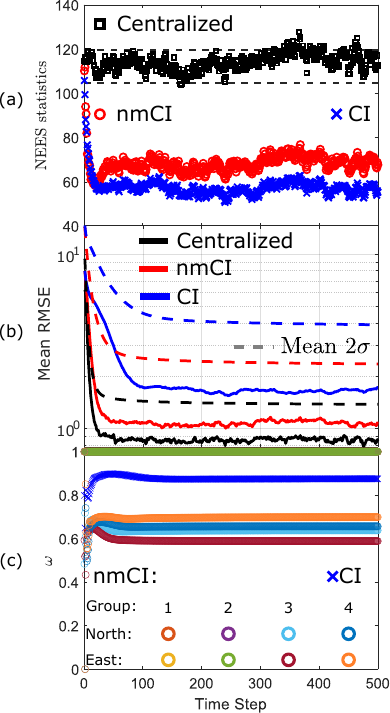}
    \caption{MC simulation results showing (a) NEES statistics, (b) mean RMSE, (c) weighing constant value $\omega$.}
    \label{fig:simResults}
    \vspace{-0.25in}
\end{figure}

The results of 15 MC runs
as computed by agent 7 are shown in Fig. \ref{fig:simResults}.
In (a) we can see the NEES chi-square consistency test \cite{rong_practical_2001}, \cite{bar-shalom_linear_2001} with 95\% confidence level for a centralized estimator (black), monolithic CI (blue), and nmCI (red). 
The results show that (i) both the CI and nmCI algorithms provide consistent results, (ii) the nmCI has a higher NEES value, which hints that the nmCI is less conservative, i.e. it is a \emph{tighter bound}.
Since, to the best of our knowledge, no quantitive measure of `less conservative' exist, we choose to compare the mean root mean squared error (RMSE) in Fig. \ref{fig:simResults}(b). 
From the graph, we can see that across all MC runs the mean RMSE and the mean average $2\sigma$ of nmCI are $33\%$ and $40\%$ smaller than CI, respectively.

Another interesting point for comparison and analysis is the $\omega$ values.
Fig. \ref{fig:simResults}(c) compares the monolithic $\omega$ value of the CI to $8$ values of the nmCI (two values for each group of robots). 
The graph shows optimal $\omega$ (minimum trace) for fusion between agents $7$ and $11$, where $\omega=1$ indicates taking agent 7's estimate, and ignoring agent $11$'s estimate. 
The dynamics of the monolithic $\omega$ vs. the non-monolithic $\omega$ (according to (\ref{eq:CI})). 
From the network topology (Fig. \ref{fig:net_topology}) we can see that agent $7$ `sits' at a much more centralized location relative to agent $11$, receiving data from different parts of the network earlier, thus its estimate is much more informative than of agent $11$. 
For the monolithic CI this results in a steady state value of $\omega=0.87$, i.e. almost ignoring agent $11$'s estimate. 
On the other hand, for the nmCI we can see that (i) for the last two agent groups (agents $9-16$), where some data flows from agent $11$ before reaching $7$, $\omega$ values range between $0.59\div 0.7$, i.e. throws away much less information from agent $11$, and (ii) for the first two agent groups (agents $1-8$), where information must pass through $7$ before reaching $11$, $\omega=1$, i.e. completely ignoring agent $11$'s estimate.

\section{Conclusion}
\label{sec:Conclusion}
In this work, we showed how to exploit the probabilistic independence structure in a multi-agent team to (i) extend the original monolithic CI to a new non-monolithic CI algorithm, and (ii) develop a conservative fusion optimization approach.
We demonstrated in a $2D$ scenario that the optimization approach converges to the nmCI solution, suggesting that it is the optimal solution.
We then test the performance of nmCI on a large-scale simulation and show that it provides $33\%$ smaller RMSE and is less conservative than CI.  

This work surfaced some interesting theoretical and practical questions that we plan to explore, such as 
(i) How to quantitatively compare the level of conservativeness, 
(ii) How can we verify that the optimization solution converged to the optimal solution, (iii) What other types of robotics applications have a similar independence structure that can be exploited, and (iv) In what scenarios enforcing the independence structure and then using nmCI might have an advantage over using monolithic CI.

\newpage
\bibliographystyle{plainnat}
\bibliography{references}

\end{document}